\pdfoutput=1

\documentclass[11pt]{article}

\usepackage{acl}

\usepackage{times}
\usepackage{latexsym}

\usepackage[T1]{fontenc}

\usepackage[utf8]{inputenc}

\usepackage{microtype}

\usepackage{inconsolata}

\usepackage{multirow}

\usepackage[ruled,vlined]{algorithm2e}
\usepackage{graphicx}
\usepackage{caption}
\usepackage{subcaption}

\newcommand{\sub}{{$\mathcal{S}$}}
\newcommand{\obj}{{$\mathcal{O}$}}
\newcommand{\rel}{{$\mathcal{R}$}}

\newcommand{\objpred}{{$\mathcal{O^{\prime}}$}}

\newcommand{\objpredindex}{{$\mathcal{O}_j^{\prime}$}}
\newcommand{\objindex}{{$\mathcal{O}_j$}}

\newcommand{\subpredindex}{{$\mathcal{S}_j^{\prime}$}}
\newcommand{\subindex}{{$\mathcal{S}_j$}}

\newcommand{\relindex}{{$\mathcal{R}_j$}}

\newcommand{\temp}[2]{$t($#1$,$#2$)$}

\newcommand{\tempindex}[2]{$t_{i}($#1$,$#2$)$}

\newcommand{\plm}[1]{$PLM($#1$)$}

\usepackage{multirow}
\usepackage{array,booktabs}

\newcolumntype {+}{ >{\global\let\currentrowstyle\relax}}
\newcolumntype {^}{ >{\currentrowstyle }}

\usepackage{xcolor}
\usepackage{booktabs}
\usepackage{soul}

\newcommand{\addition}[1]{{#1}}

\newif\ifshowChanges

\showChangestrue 
\showChangesfalse 

\ifshowChanges
    \newcommand{\CR}[1]{{\color{blue}#1}}
    \newcommand{\CRR}[1]{{\st{#1}}}

\else
    \newcommand{\CR}[1]{{#1}}
    \newcommand{\CRR}[1]{\ignorespaces}
    
\fi

\title{The Queen of England is not England's Queen: \\On the Lack of Factual Coherency in PLMs}

\author{Paul Youssef\textsuperscript{$\dag$}  Jörg Schlötterer\textsuperscript{$\dag\ddag$} Christin Seifert\textsuperscript{$\dag$}\\
 \textsuperscript{$\dag$}University of Marburg, \textsuperscript{$\ddag$}University of Mannheim\\
  \texttt{\{paul.youssef, joerg.schloetterer, christin.seifert\}@uni-marburg.de}\\}

\begin{document}
\maketitle
\begin{abstract}
Factual knowledge encoded in Pre-trained Language Models (PLMs) enriches their representations and justifies their use as knowledge bases. Previous work has focused on probing PLMs for factual knowledge by measuring how often they can correctly predict an \emph{object} entity given a subject and a relation, and improving fact retrieval by optimizing the prompts used for querying PLMs. In this work, we consider a complementary aspect, namely the coherency of factual knowledge in PLMs, i.e., how often can PLMs predict the \emph{subject} entity given its initial prediction of the object entity. This goes beyond evaluating how much PLMs know, and focuses on the internal state of knowledge inside them. Our results indicate that PLMs have low coherency using manually written, optimized and paraphrased prompts, but including an evidence paragraph leads to substantial improvement. This shows that PLMs \CR{fail to model inverse relations and} need further enhancements to be able to handle retrieving facts from their parameters in a coherent manner, and to be considered as knowledge bases.  

\end{abstract}

\section{Introduction}

Pre-trained Language Models (PLMs) are probed for factual knowledge to investigate their usage as knowledge bases, and gain a better understanding of the rich representations they provide~\cite{petroni-etal-2019-language}. 
Previous extensions have focused on extracting more facts~\cite{zhong-etal-2021-factual, li-etal-2022-spe}, increasing the consistency of PLMs to paraphrased prompts~\cite{elazar-etal-2021-measuring}, identifying the parts of PLMs that are responsible for storing knowledge~\cite{dai-etal-2022-knowledge} and updating facts in them~\cite{meng-etal-2022-locating, meng-etal-2023-massediting}. 

\begin{figure}[ht]

\includegraphics[width=\columnwidth, trim={5.4cm 21cm 0.5cm 4.3cm}, clip]{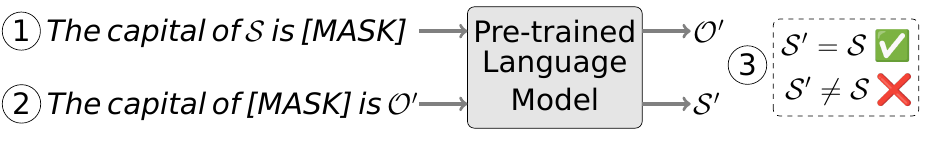}

\caption{\CR{Probing} coherency in PLMs. 1) The PLM makes a prediction based on an entity $\mathcal{S}$ and a relation. 2) The PLM makes a second prediction based on the same relation and its first prediction $\mathcal{O}^{\prime}$. 3) If the PLM predicts $\mathcal{S}$ in the second step it shows coherent behavior.}
\label{fig:overview}
\end{figure}
\CR{More recently, \citet{berglund-etal-2023-reversal} study the generalization abilities of PLMs from ``A is B'' to ``B is A'', and show that if a PLM is trained on ``The capital of Malta is Valetta'' it will not be able to \emph{correctly} answer the question: ``Which country has Valetta as its capital?''.} In this work, we introduce a\CR{n intrinsic and } complementary aspect, namely the \emph{coherency} of PLMs with respect to factual knowledge. \CR{Coherency is not concerned with correctness of the PLMs' predictions, but with the internal state of knowledge in PLMs and its consistency. More concretely, we first ask a PLM to answer the question: ``What is the capital of Malta?'', and if it answers ``Berlin'', we ask it to answer the question: ``Which country has Berlin as its capital?'', and if it answers ``Malta'', then we say that the PLM has answered coherently (even though the answer is factually wrong). Note that in practice we use Cloze prompts instead of questions to make the task closer to language modeling (see Figure~\ref{fig:overview}).} Intuitively, if a human can \CRR{correctly} tell the capital of a country given that country's name, then she is also able to tell the country given its capital's name. Given that PLMs are queried with a subject and a relation to extract a object, we define coherency as the ability of the PLM to infer the subject given its initial prediction for the object entity and vice versa. 


Our contributions are the following: (1) we introduce coherency to investigate the internal state of factual knowledge in PLMs; (2) we evaluate different PLMs, showing that they have low coherency; (3) we show that optimized and paraphrased prompts do not improve coherency, but the use of evidence paragraphs substantially improves coherency. We make our code available.\footnote{\url{https://github.com/paulyoussef/coherency}} 


\section{Coherency}
PLMs are known to capture vast amount of facts from their pre-training corpora. This has encouraged the community to consider using them as knowledge bases (KBs)~\cite{petroni-etal-2019-language, sung-etal-2021-language}, which can be constructed without expensive annotations, and which can easily be queried using natural language. However, the use of PLMs as KBs has many limitations~\cite{alkhamissi-etal-2022-review}. For example, PLMs are quite sensitive to their prompts, and cannot be easily updated with new facts. Factual knowledge in PLMs is estimated by evaluating how often PLMs can correctly predict an object entity \obj{}, given a subject entity \sub{} and a relation \rel, when provided with a prompt which contains the subject and the relation: \temp{\sub}{\rel}, where $t$ is a function that maps a subject entity and a relation to a prompt in natural language that contains the given entity and expresses the relation in natural language, e.g., \texttt{(Malta, capital-of)} $\rightarrow$ \texttt{``The capital of Malta is [MASK]''}. In this work, we focus on evaluating the coherency of PLMs with respect to the factual knowledge stored in their parameters, i.e., how often can PLMs predict \sub{} using \temp{\objpred}{\rel}, given that it predicted \objpred{} using~\temp{\sub}{\rel}. For example, \texttt{``The capital of [MASK] is \CR{Berlin}\CRR{Valetta}''} $\rightarrow$ \texttt{Malta} is coherent with \texttt{``The capital of Malta is [MASK]''} $\rightarrow$ \texttt{``\CR{Berlin}\CRR{Valetta}''}. We do not evaluate if the predictions are factually correct, because we are interested in the coherency of the PLMs' world view, regardless of its correctness \addition{.We show and discuss correctness scores in Appendix~\ref{app:add-results}.} 

Coherency can be easily calculated for 1-1 relations, but is more challenging, if we consider N-1 or N-M relations, since multiple entities could be correct when trying to predict the subject entity. To address this, we exclude all correct entities except the ground truth subject \sub{} in the second inference step, following~\citet{bordes-etal-2013-translating} \CR{and~\citet{petroni-etal-2019-language}}. Since PLMs are known to have certain biases and are sensitive to the prompts, we start with predicting the object given the subject in a first round. In a second round, we start by predicting the subject given the object. The complete algorithm for estimating coherency in PLMs for all types of relations is shown in Algorithm \ref{alg:coherency}. After estimating coherency for each relation, we macro-average over all relations, because we are interested in the average performance for the use case of PLMs-as-KBs, which involves storing facts from different types of relations.

\begin{algorithm}[h] 
  \caption{Coherency in PLMs}
  \label{alg:coherency}
\KwIn{PLM, dataset with $n$ relations}
\KwOut{coherency}

scores\_per\_relation = [] \\
\tcp{iterate over relations}
\For{$i \gets 1$ \KwTo $n$}{
scores = []\\
\tcp{iterate over instances}
 \For{$j \gets 1$ \KwTo $m$}{
    \tcp{round 1: start with object}
    \objpredindex{} = \plm{\tempindex{\subindex}{\relindex}}\\
    exclude correct answers except \subindex{}\\
    \subpredindex{} = \plm{\tempindex{\objpredindex{}}{\relindex}}\\
    \If{ partial\_match(\subpredindex{}, \subindex{})}{
        scores.append(1)
    }
    \Else{
        scores.append(0)
    }
    
    \tcp{round 2: start with subject}
    \subpredindex{} = \plm{\tempindex{\objindex}{\relindex}}\\
    exclude correct answers except \objindex{}\\
    \objpredindex{} = \plm{\tempindex{\subpredindex}{\relindex}}\\
    \If{ partial\_match(\objpredindex{}, \objindex{} )}{
        scores.append(1)
    }
    \Else{
        scores.append(0)
    }
    }
    scores\_per\_relation.append(mean(scores))
  }
\KwRet{mean(scores\_per\_relation )}
\end{algorithm}

\section{Experimental Setup}
Here, we describe the data and PLMs, which we use, and our experiments in detail.

\subsection{Data}
In our experiments, we use the T-REx~\cite{elsahar-etal-2018-rex} subset of LAMA~\cite{petroni-etal-2019-language}, which is often used to estimate factual knowledge in PLMs. T-REx consists of 41 relations with their corresponding templates, and subject-object pairs, for which the relations hold in English. For each of the relations, a manually-written template is provided, which we use to construct the prompts. Some statistics and an example from the T-REx subset are shown in Table~\ref{tab:lama} in Appendix~\ref{app:datasets}.

\subsection{How coherent are PLMs?}
\label{exp1}
In this experiment, we aim to find out how coherent are PLMs. We \CR{mostly focus on} PLMs which are trained to fill in the blanks based on context, since \CR{these make use of a bidirectional context, and we expect them to perform better than autoregressive PLMs on this task}\CRR{auto-regressive PLMs can only complete text in one direction}. More specifically, we consider BERT~\cite{devlin-etal-2019-bert}, InformBERT~\cite{sadeq-etal-2022-informask}, T5~\cite{raffel-etal-2020-t5}, and T5-SSM~\cite{guu-etal-2020-realm, roberts-etal-2020-much}. InformBERT adapts the masking strategy of BERT to focus on more informative tokens. T5-SSM models are additionally trained with Salient Span Masking objective (SSM), which masks only named entities in the pre-training phase. More information about the models are provided in Appendix~\ref{app:mlm}. If available, we consider several sizes of the same model in order to investigate the effect of scaling PLMs on coherency. For BERT-based models, we only consider entities that correspond to one token, in order to adhere to the task format from pre-training. We evaluate all models in a zero-shot setting with no finetuning, since we are interested in the coherency of factual knowledge in PLMs after the pre-training phase. For BERT-based models, we choose the token with the highest probability. For T5-models, we use beam search with 10 beams. We use partial match, which returns true if one of the two predictions is contained in the other after converting both to lower case, when comparing the predictions against the ground truth entities. 

\CR{For completeness, we also evaluate on autoregressive PLMs. More specifically, we consider GPT-2~\cite{radford-etal-2019-language} and GPT-Neo~\cite{gao-etal-2020-pile, gpt-neo}. For autoregressive PLMs, we use typed querying~\cite{kassner-etal-2021-multilingual}, i.e., we extract a probability distribution over a pre-defined set of entities from the model, and choose the most probable entity as the final prediction. Typed querying makes it easy to extract valid answers (entities) from the PLMs' outputs, but also makes the task easier for PLMs since it restricts the output space. We extend the templates from LAMA such that the subject/object entities appear at the very end. We consider autoregressive PLMs only in this experiment.} 

\subsection{Do optimized prompts improve coherency?}
Optimizing prompts leads to better fact retrieval~\cite{zhong-etal-2021-factual}. In this experiment, we investigate whether optimized prompts lead to higher coherency as well. We utilize~\citet{shin-etal-2020-autoprompt}' optimized prompts for T-REx. These prompts differ from one model to another, and from the models we consider, optimized prompts are only available for BERT models.

\subsection{Does providing an evidence paragraph increase coherency?}
PLMs can fill in the blanks based on the knowledge they have stored in their parameters (parametric knowledge), or based on information that is provided in their inputs (contextual knowledge). The latter boils down to extracting the right information from the input. Previous work has shown that providing evidence paragraphs as additional inputs makes PLMs' predictions more factual~\cite{petroni-etal-2020-context}. Here, we investigate how these evidence paragraphs affect the coherency of factual knowledge in PLMs. The provided evidence paragraphs from LAMA contain a Wikipedia paragraph that express\CR{es} the facts. We append the evidence paragraphs to the inputs from the first experiment. 

\subsection{Is Coherency stable across paraphrased prompts?}
PLMs are known to be sensitive to the provided prompts, i.e., small insignificant changes, that preserve the meaning cause the PLMs to change their predictions~\cite{elazar-etal-2021-measuring}. As a result, retrieving facts from PLMs is highly affected by the prompts used. In this experiment, we consider the effect of using paraphrased prompts on coherency. Does coherency stay the same across different prompts or is it highly variant? We evaluate whether coherency varies with paraphrased prompts from~\citet{elazar-etal-2021-measuring}'s ParaRel dataset. ParaRel provides paraphrases for 38 of the 41 relations in T-Rex. For each one of the 38 relations, we randomly select a template from ParaRel, and measure how coherency is changed over 10 runs. We consider bert-base and t5-base for this experiment.

\section{Results and Discussion}
\label{sec:discussion}
The results for the first three experiments are shown in Table~\ref{tab:results}. \CR{We show the results for autoregressive PLMs separtely in Table~\ref{tab:arlms}, because we probe autoregressive PLMs with typed querying}. We do not evaluate if the predictions are factually correct. For correctness scores see Table~\ref{tab:results_correct} in Appendix~\ref{app:add-results}. Since we considered only one-token entities from T-REx for BERT models, we show a normalized version of the results on this subset for better comparability in Table~\ref{tab:results_normalized}, and the results with the total number of instances in Table~\ref{tab:results_w_instances} in Appendix~\ref{app:add-results}.

\paragraph{PLMs show poor coherency.} \CR{We notice that all PLMs have poor coherency. Autoregressive PLMs perform even worse than masked PLMs, even though the task is made easier for them through typed querying (cf. Section~\ref{exp1}). The poor performance of autoregressive PLMs might be due to their unidirectional training objective, whereas masked PLMs make use of a bidirectional context.}
\CRR{We notice that all \emph{masked PLMs} have poor coherency scores. This complements berglund-etal-2023-reversal's findings that \emph{autoregressive PLMs} trained on ``A is B'' fail to generalize to ``B is A''.} Increasing the number of parameters in T5 models leads to consistent improvements in performance. However, this does not generalize to the BERT models (bert-base performs better than bert-large), and to the T5 models that are trained with SSM (t5-large-ssm performs better than t5-3b-ssm). The SSM objective is beneficial for the large variant of T5 (t5-large-ssm improves by 6.5 percentage points over t5-large, and even outperforms t5-3b, which has 4 times as many parameters). Contrarily, this improvement does not generalize to the 3b variant (t5-3b outperforms its SSM counterpart). InformBERT falls short of normal BERT, even though it was shown to outperform BERT, when it comes to facts retrieval~\cite{sadeq-etal-2022-informask}. Hence, better facts retrieval does not necessarily affect coherency positively. In general, scaling and entity-centric training objectives have to some extent a positive effect on coherency. \addition{We also notice that in most cases models perform worse in the first round. Round 1 can be more difficult, since it may involve predicting a specific subject based on a generic object in the second step (e.g.,\texttt{``[MASK] is located in Bern''}), whereas the second round goes into opposite and easier direction (\texttt{``University of Bern is located in [MASK]''}). PLMs are known to not provide specific answers~\cite{huang-etal-2023-language}.} 

\CR{We show the results per relation type in Table~\ref{tab:results_per_type} in Appendix~\ref{app:add-results}. The evaluation dataset contains 2 \textbf{1-1} relations , 23 \textbf{N-1} relations and 16 \textbf{N-M} relations with 3 of the 16 \textbf{N-M} relations being symmetric. Most PLMs have high coherency on \textbf{1-1} relations, but the number of instances for these relations is limited (747 at most), on \textbf{N-1}, \textbf{N-M} and symmetric relations the performance drops significantly. This shows that \textbf{N-1} and \textbf{N-M} relations are challenging for PLMs not just with respect to facts retrieval~\cite{petroni-etal-2019-language}, but also with respect to developing a coherent knowledge state.}

\CR{We also show and categorize examples from different PLMs in Table~\ref{tab:examples} in Appendix~\ref{app:examples}. In general, one can notice that incoherent predictions are due to : 1) The answer being incorrect in the first step, making it more difficult to predict the answer in the second step (rows 6-7); 2) The templates being not specific enough allowing for non-factual completions (row 8); 3) missing context to retrieve correct relation for non 1-1 relations (row 3).}


\paragraph{Optimizing prompts does not help.} Optimized prompts lead to a drop in coherency in the second experiment (see results under optimized prompts in Table~\ref{tab:results}) l. This shows that prompts that better retrieve object entities does not help retrieve the corresponding subject entities. Previous work showed that optimized prompts overfit the facts distribution of objects~\cite{cao-etal-2021-knowledgeable}, which might negatively affect their ability to retrieve the subject entities. \addition{This is also evident by the difference in scores between the two rounds.} 

\paragraph{Evidence paragraphs improve coherency.} Including evidence paragraphs in the inputs substantially improves performance (see results under evidence paragraphs in Table~\ref{tab:results}). This shows that PLMs are better at extracting answers from their inputs than recalling them from their parameters. In fact, adding an evidence paragraph reduces the performance gaps among models of different sizes and pre-training objectives. This suggests that retrieval-based approaches are indeed a promising alternative to scaling language models~\cite{kandpal-etal-2023-large}. Still, coherency is not high under this setting as well. We believe this is due to the PLMs failing to extract the correct entities or to the conflicts between contextual and parametric knowledge in PLMs~\cite{neeman-etal-2023-disentqa}.

\paragraph{Coherency varies across paraphrases.} Table~\ref{tab:results_para} shows the minimum, average and maximum coherency scores with paraphrased prompts. A break-down in relations is available in Appendix~\ref{app:add-results} (Fig.~\ref{fig:avg}).\footnote{Note that, for this experiment, we use only 38 of the 41 relations in T-Rex -- The ones for which paraphrases exist.} As with fact retrieval, the results indicate that prompts have a significant effect on the performance. For example, there are more than 25 percentage points difference in coherency between the min and max scores for t5-base. Still, even when considering the best prompts, the overall coherency score is low. 

 
\CR{In general, our analysis shows that PLMs do not possess a coherent knowledge state. The low coherency might be due: 1) The fact that PLMs make predictions based on shallow surface level features~\cite{poerner-etal-2020-e, li-etal-2022-pre}, which makes PLMs output relevant but incoherent and non-factual predictions (for an example see row 6 in Table~\ref{tab:examples}). This is inherent to all PLMs, and requires further architectural improvements; 2) The training data for PLMs, which might be biased towards certain entities (the more frequent ones); 3) The uni-directional training in the case of autoregressive PLMs that makes PLMs sensitive to the order in which the entities are observed.}

\begin{table}[ht]
\small
\renewcommand*{\arraystretch}{1.1}
\centering
\begin{tabular}{lrrr}
\toprule
\textbf{Model} &  \textbf{Round 1} &  \textbf{Round 2} &  \textbf{Avg.} \\
\midrule
bert-base-uncased   &          9.74 &         11.81 &      10.78 \\
bert-large-uncased  &          9.83 &         10.29 &      10.06 \\
InformBERT          &          8.04 &         11.55 &       9.79 \\
t5-base             &          9.02 &         10.29 &       9.66 \\
t5-large            &          9.07 &         12.03 &      10.55 \\
t5-3b               &          8.62 &         23.90 &      16.26 \\
t5-large-ssm        & \textbf{9.89} &         \textbf{24.23} &  \textbf{17.06} \\
t5-3b-ssm           &          8.97 &         20.88 &      14.92 \\
\hline
\multicolumn{3}{l}{\textbf{w/ optimized prompts}}  \\
\hline
bert-base-uncased  &          1.52 &         \textbf{12.80} & \textbf{7.16} \\
bert-large-uncased &  \textbf{1.87} &        7.38 &       4.62 \\
\hline
\multicolumn{3}{l}{\textbf{w/ evidence paragraphs}}  \\
\hline
bert-base-uncased   &         22.30 &         39.87 &      31.09 \\
bert-large-uncased  &         21.05 &         41.98 &      31.52 \\
InformBERT          &         43.07 &         46.40 &      44.74 \\
t5-base             &         41.40 &         58.31 &      \underline{\textbf{49.85}} \\
t5-large            &         31.46 &         55.15 &      43.31 \\
t5-3b               &         27.06 & \textbf{62.89} &      44.98 \\
t5-large-ssm        & \textbf{50.17} &         43.97 &      47.07 \\
t5-3b-ssm           &         48.52 &         41.81 &      45.17 \\

\bottomrule
\end{tabular}

\caption{Coherency score per round and on average for different PLMs using manually-written, optimized prompts and evidence paragraphs. The highest performance under each category is in \textbf{bold}, and the best performance overall is \underline{underlined}.}
\label{tab:results}
\end{table}

\begin{table}
\renewcommand*{\arraystretch}{1.1}
\small
\centering
\begin{tabular}{lrrr}
\toprule
\textbf{Model} &  \textbf{Round 1} &  \textbf{Round 2} &  \textbf{Avg.} \\
         
\midrule
gpt2         &          0.24 &          3.98 &       2.11 \\
gpt-neo-1.3B &          0.44 &         \textbf{12.85} &       \textbf{6.65} \\
gpt-neo-2.7B &          \textbf{0.56} &         11.82 &       6.19 \\
\bottomrule
\end{tabular}
\caption{\CR{Coherency score per round and on average for autoregressive PLMs using manually-written prompts. The highest performance is in \textbf{bold}. Autoregressive PLMs are probed using typed querying.}}
\label{tab:arlms}
\end{table}

\begin{table}
\renewcommand*{\arraystretch}{1.1}

\centering
\small
\begin{tabular}{lllll}
\hline
\textbf{Model} & \textbf{Min.} & \textbf{Avg.} & \textbf{Max.} & \textbf{\#Instances} \\
\hline
bert-base-uncased & 3.74 & 11.16  & 19.25 & 2852 \\
          t5-base & 6.51 & 16.88 & 31.69 &  27788 \\

\hline
\end{tabular}
\caption{Coherency scores with different paraphrases. We show the results with the worst/average/best performing prompts per relation.}
\label{tab:results_para}
\end{table}

\section{Related Work}

\paragraph{Reversal curse.} \CR{\citet{berglund-etal-2023-reversal} investigate the generalization abilities of autoregressive PLMs from one data form, that is encountered during training (A is B), to another (B is A), showing a generalization failure. \citet{berglund-etal-2023-reversal} refer to this generalization inability in autoregressive PLMs as the \emph{reversal curse}. Our work is close but complementary to theirs. We focus on the coherency of the internal state of factual knowledge in autoregressive  \emph{and} masked PLMs, \emph{regardless} of how correct the PLMs' predictions are.} 

\paragraph{Factual knowledge in PLMs.} PLMs contain vast amounts of linguistic~\cite{tenney-etal-2018-context, jawahar-etal-2019-bert}, commonsense~\cite{davison-etal-2019-commonsense} and factual knowledge~\cite{roberts-etal-2020-much} that is captured during pre-training. Many works focus on factual knowledge in PLMs~\cite{youssef-etal-2023-give}, since factual knowledge is said to contribute to the rich presentations produced by PLMs, and potentially justifies the use of PLMs as KBs~\cite{petroni-etal-2019-language, ye-etal-2022-zerogen}. For example, \citet{shin-etal-2020-autoprompt, zhong-etal-2021-factual} optimize prompts to extract more facts from PLMs, ~\citet{elazar-etal-2021-measuring, fierro-sogaard-2022-factual} investigate the sensitivity of PLMs to paraphrased prompts, \cite{malkin-etal-2022-coherence, wang-etal-2023-towards-alleviating} debias the outputs of PLMs for better facts extraction, \citet{meng-etal-2022-locating, meng-etal-2023-massediting} address editing facts in PLMs to make it possible to correct and update facts. However, these works collectively focus on extrinsic aspects. We focus on a more intrinsic aspect, i.e., the coherency of factual knowledge inside PLMs. This complements aspects addressed in previous work.

\section{Conclusion}
In this work, we focused on evaluating the coherency of factual knowledge in PLMs. We considered the use of manually-written, optimized, and paraphrased prompts. Our results indicate poor coherency. The inclusion of an evidence paragraph leads to substantial improvements. This shows that PLMs can leverage contextual knowledge better than parametric knowledge \CR{and highlights the importance of retrieval-augmented PLMs}. We believe that further improvements are needed to improve coherency in PLMs, and to consider them as alternatives to KBs. \CR{We believe that future work should focus on further improving PLMs on the architectural level, the data level, and the interface between them (pre-training objectives).} 

\clearpage

\clearpage

\section{Limitations}
\label{sec:limitations}
Coherency can be easily determined using 1-1 relations. For N-1 or N-M relations, some potential answers should be excluded. However, it is quite difficult to exclude every possible answer for certain relations (e.g., everyone who is an English native speaker) from the model's vocabulary. We only excluded answers that are present in LAMA, following previous work~\cite{bordes-etal-2013-translating} and \cite{petroni-etal-2019-language}. This might have had a negative effect on the results (cf. Section~\ref{sec:discussion}, discussion of lower scores in round 1).

\section*{Acknowledgement}
We thank Alessandro Noli for helpful discussions.   

\bibliography{anthology,custom}

\appendix

\section{Additional Results}
\label{app:add-results}

\addition{\paragraph{Correctness.} We investigate how correct the PLMs' predictions are. For each instance, we count how often the first prediction in the first round (\textbf{c1}), and in the second round (\textbf{c2}) were correct. We only consider the first predictions in each round, since having the incorrect answer in the first inference step of each round makes it more difficult for the model to answer correctly in the second inference step. We also report how often are all predictions correct (\textbf{all correct}). We calculate each score for all relations, and average over all relations. Results are shown in Table~\ref{tab:results_correct}. We notice that for all models the \textbf{c1} scores are higher than the \textbf{c2} scores. We believe this is because in the first inference step in round 1, models predict \emph{object} entities, whereas in the first step of round 2 they predict \emph{subject} entities. Predicting subject entities is more difficult, since their corresponding mask tokens are placed at the beginning of the templates. This allows for valid completions that do not contain any entities. For example, if the template is \texttt{``[MASK] is the capital of Malta''}, then \texttt{``It''} is also a valid completion with no entities. Additionally, predicting the subject entity based on the object entity might be ambiguous (see discussion in Section~\ref{sec:discussion}). }



\paragraph{Coherency scores per relation type.} Coherency scores per relation type are shown in Table~\ref{tab:results_per_type}.

\paragraph{Coherency on a subset.}Table~\ref{tab:results_normalized} shows a normalized version of the coherency scores using manually-written prompts.

\begin{figure*}
    \centering

    \begin{subfigure}[b]{\textwidth}
        \includegraphics[width=\textwidth]{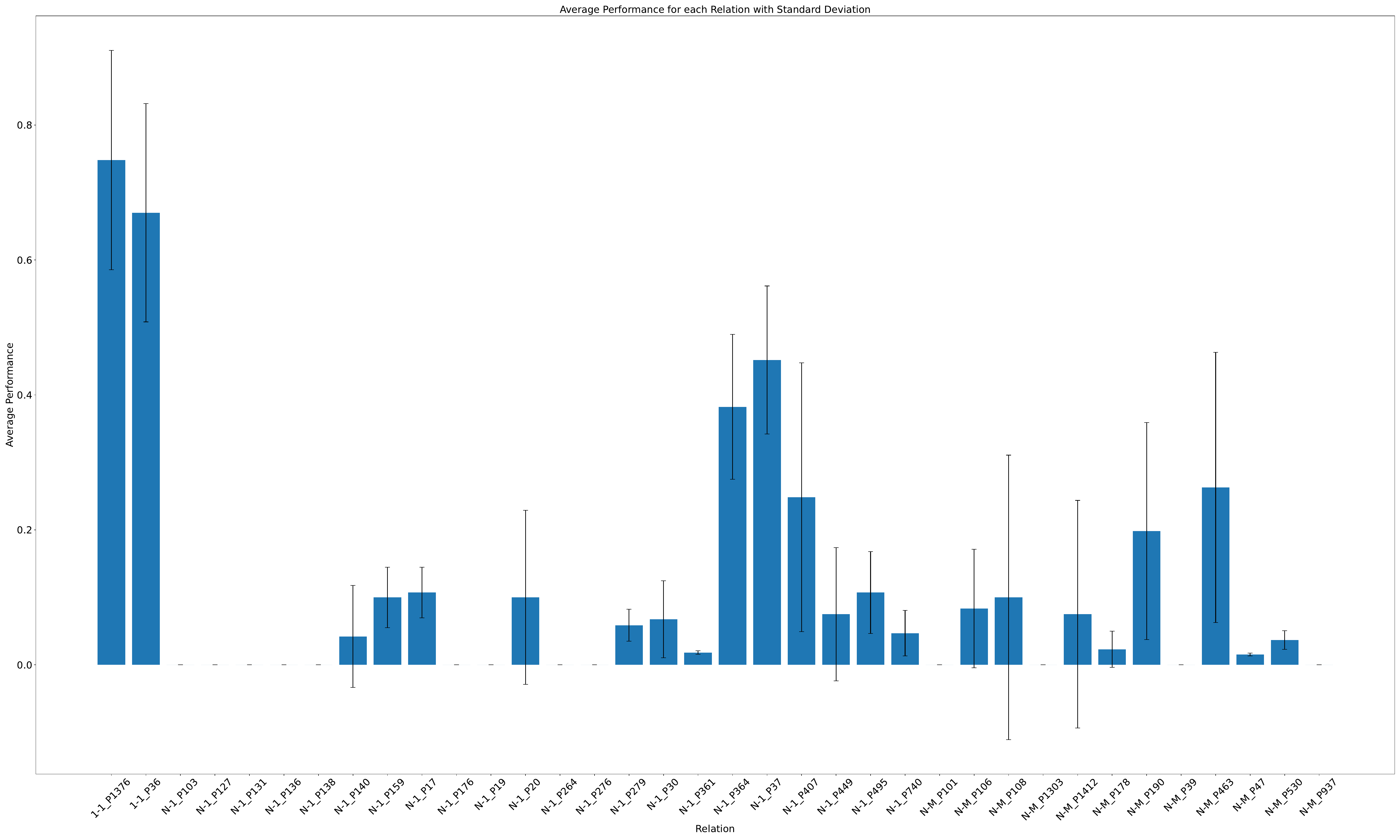}
        \caption{bert-base-uncased}
        \label{fig:subfig1}
    \end{subfigure}

    \begin{subfigure}[b]{\textwidth}
        \includegraphics[width=\textwidth]{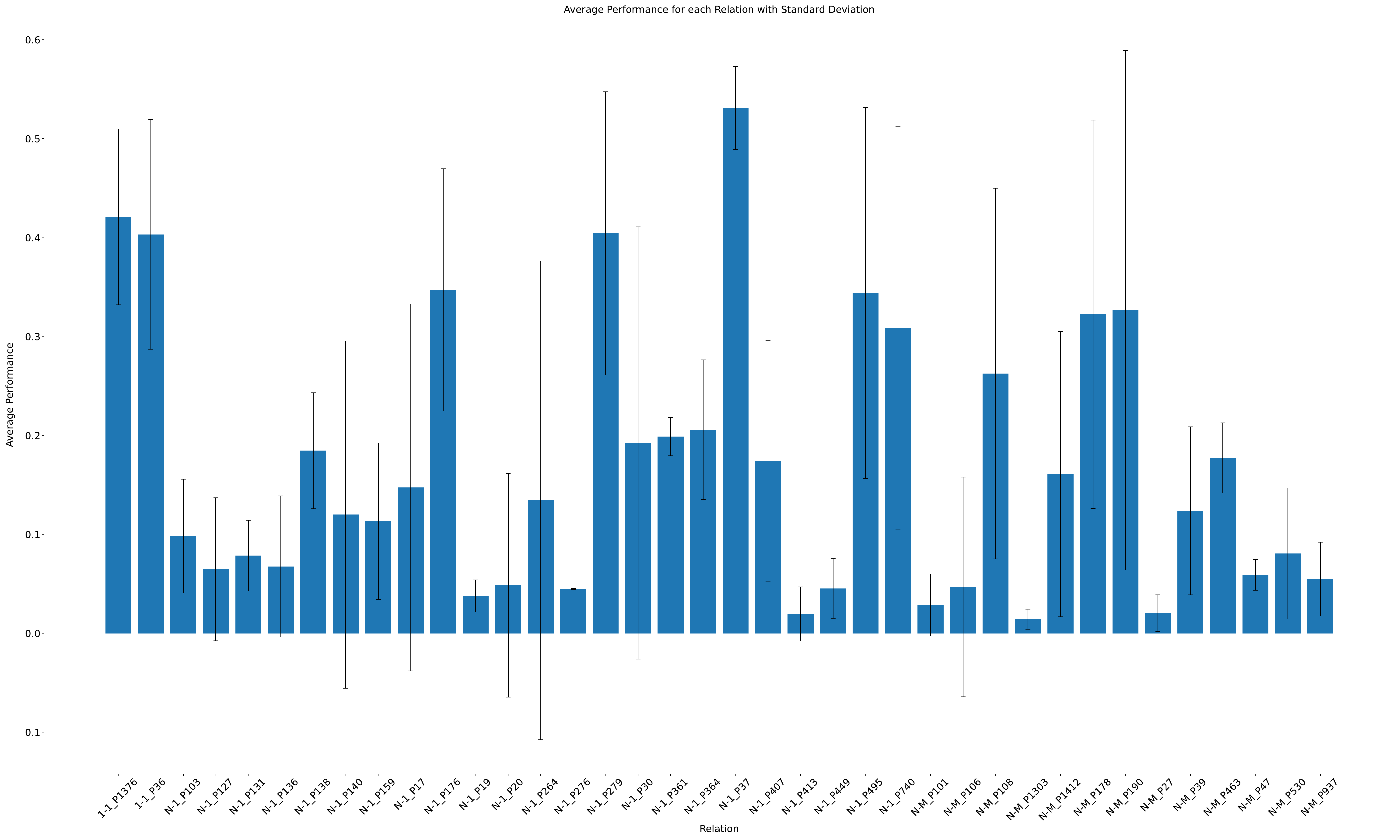}
        \caption{t5-base}
        \label{fig:subfig2}
    \end{subfigure}

    \caption{Average coherency with standard deviation when using paraphrased prompts over different relations. }
    \label{fig:avg}
\end{figure*}

\paragraph{Coherency over relations with different paraphrases.} Figure~\ref{fig:avg} shows the average coherency scores with standard deviation over different relations when using paraphrased prompts. Note that bert-base-uncased has less relations than t5-base (36 vs. 38), since some relations ended up with no instances after excluding multi-token entities. In general, we notice high standard deviation for most relations.

\paragraph{Coherency scores with the number of instances.} Table~\ref{tab:results_w_instances} shows the coherency scores with the size of the test set in instances.

\begin{table*}
\centering
\begin{tabular}{lccccc}
\toprule
\textbf{Model} &  \textbf{c1} &  \textbf{c2} & \textbf{All correct}&  \textbf{\#relations} &  \textbf{\#Instances} \\
\midrule
bert-base-uncased   &   30.77 &    8.55 &      4.27 &        39 &        2919 \\
bert-large-uncased  &   25.96 &    8.39 &      4.22 &        39 &        2919 \\
InformBERT          &   22.33 &    5.97 &      4.34 &        39 &        2926 \\
t5-base             &   11.03 &    6.21 &      1.30 &        41 &       29672 \\
t5-large            &   14.77 &    6.26 &      1.70 &        41 &       29672 \\
t5-3b               &   20.93 &    6.10 &      2.33 &        41 &       29672 \\
t5-large-ssm &   18.42 &    4.69 &      2.73 &        41 &       29672 \\
t5-3b-ssm    &   19.61 &    4.28 &      2.96 &        41 &       29672 \\

\hline
\multicolumn{6}{c}{Autoregressive PLMs}  \\
\hline
gpt2                    &    7.70 &    0.43 &      0.04 &        41 &       29672 \\
gpt-neo-1.3B &   17.65 &    0.93 &      0.13 &        41 &       29672 \\
gpt-neo-2.7B &   18.50 &    1.31 &      0.22 &        41 &       29672 \\
\hline
\multicolumn{3}{l}{\textbf{w/ optimized prompts}}  \\
\hline
bert-base-uncased  &   25.27 &    1.49 &      0.02 &        39 &        2919 \\
bert-large-uncased &   31.92 &    2.94 &      0.10 &        39 &        2919 \\
\hline
\multicolumn{3}{l}{\textbf{w/ evidence paragraphs}}  \\
\hline
bert-base-uncased   &   46.98 &   19.97 &     11.12 &        39 &        2919 \\
bert-large-uncased  &   49.66 &   20.27 &     12.98 &        39 &        2919 \\
InformBERT   &   49.42 &   45.92 &     24.95 &        39 &        2926 \\
t5-base             &   59.77 &   39.28 &     23.99 &        41 &       29672 \\
t5-large            &   59.31 &   27.57 &     15.77 &        41 &       29672 \\
t5-3b               &   57.35 &   23.17 &     11.73 &        41 &       29672 \\
t5-large-ssm &   44.47 &   47.61 &     23.10 &        41 &       29672 \\
t5-3b-ssm    &   41.44 &   46.40 &     21.41 &        41 &       29672 \\

\bottomrule
\end{tabular}

 \caption{\addition{Correctness scores in the first inference step of the first round (\textbf{c1}), the second round (\textbf{c2}), and in all inference steps (\textbf{all correct}). Results are averaged over all relations. BERT-based models have less relations and instances, because we consider only one-token entities for these models.}}
\label{tab:results_correct}
\end{table*}

\begin{table*}
\renewcommand*{\arraystretch}{1.1}
\centering
\tiny
\begin{tabular}{lrrrrrrrrrr}
\toprule
\textbf{Relation Type} & \multicolumn{2}{c}{\textbf{1-1}} & \multicolumn{2}{c}{\textbf{N-1}} & \multicolumn{2}{c}{\textbf{N-M}}& \multicolumn{2}{c}{\textbf{symmetric}} & \multicolumn{2}{c}{\textbf{All}} \\ \midrule
\textbf{Model}  & \textbf{Coherency} & \textbf{\#Instances} & \textbf{Coherency} & \textbf{\#Instances} & \textbf{Coherency} & \textbf{\#Instances} &    \textbf{Coherency} & \textbf{\#Instances} & \textbf{Coherency} & \textbf{\#Instances} \\
\midrule
bert-base-uncased             &  84.11 &        232 &   5.93 &        633 &   8.10 &       2054 &     12.57 &       1927 &  10.78 &       2919 \\
bert-large-uncased            &  82.71 &        232 &   6.65 &        633 &   5.38 &       2054 &     15.10 &       1927 &  10.06 &       2919 \\
InformBERT                    &  81.03 &        232 &   5.28 &        637 &   6.91 &       2057 &     18.46 &       1929 &   9.79 &       2926 \\
t5-base                       &  36.84 &        747 &   8.55 &      16838 &   7.84 &      12087 &      8.61 &       2882 &   9.66 &      29672 \\
t5-large                      &  48.90 &        747 &   6.90 &      16838 &  11.02 &      12087 &     14.87 &       2882 &  10.55 &      29672 \\
t5-3b                         &  61.21 &        747 &  14.84 &      16838 &  12.68 &      12087 &     21.41 &       2882 &  16.26 &      29672 \\
t5-large-ssm                  &  75.96 &        747 &  17.22 &      16838 &   9.46 &      12087 &      7.44 &       2882 &  17.06 &      29672 \\
t5-3b-ssm                     &  76.36 &        747 &  13.94 &      16838 &   8.66 &      12087 &     13.00 &       2882 &  14.92 &      29672 \\
\hline
\multicolumn{10}{c}{Autoregressive PLMs}  \\
\hline
gpt2                          &   0.26 &        747 &   1.46 &      16838 &   3.27 &      12087 &      0.16 &       2882 &   2.11 &      29672 \\
gpt-neo-1.3B                  &   3.40 &        747 &   9.71 &      16838 &   2.65 &      12087 &      0.19 &       2882 &   6.65 &      29672 \\
gpt-neo-2.7B                  &   4.59 &        747 &   6.37 &      16838 &   6.14 &      12087 &      0.51 &       2882 &   6.19 &      29672 \\
\hline
\multicolumn{10}{l}{\textbf{w/ optimized prompts}}  \\
\hline

bert-base-uncased  &   1.46 &        232 &   7.54 &        633 &   7.35 &       2054 &      2.36 &       1927 &   7.16 &       2919 \\
bert-large-uncased &   2.38 &        232 &   6.85 &        633 &   1.66 &       2054 &      7.23 &       1927 &   4.62 &       2919 \\
\hline
\multicolumn{10}{l}{\textbf{w/ evidence paragraphs}}  \\
\hline

bert-base-uncased   &  87.78 &        232 &  26.49 &        633 &  30.27 &       2054 &     22.66 &       1927 &  31.09 &       2919 \\
bert-large-uncased  &  90.30 &        232 &  28.13 &        633 &  28.65 &       2054 &     26.61 &       1927 &  31.52 &       2919 \\
InformBERT          &  84.06 &        232 &  42.05 &        637 &  43.43 &       2057 &     31.86 &       1929 &  44.74 &       2926 \\
t5-large-ssm        &  84.73 &        747 &  46.38 &      16838 &  43.35 &      12087 &     32.50 &       2882 &  47.07 &      29672 \\
t5-3b-ssm           &  86.95 &        747 &  44.80 &      16838 &  40.47 &      12087 &     27.10 &       2882 &  45.17 &      29672 \\
t5-base             &  73.86 &        747 &  49.91 &      16838 &  46.77 &      12087 &     30.37 &       2882 &  49.85 &      29672 \\
t5-large            &  66.94 &        747 &  42.65 &      16838 &  41.30 &      12087 &     26.10 &       2882 &  43.31 &      29672 \\
t5-3b               &  74.16 &        747 &  45.32 &      16838 &  40.84 &      12087 &     26.93 &       2882 &  44.98 &      29672 \\

\bottomrule
\end{tabular}
\caption{Coherency scores per relation type.}
\label{tab:results_per_type}
\end{table*}

\begin{table}
\centering
\begin{tabular}{lcc}
\hline 
\textbf{Model} & \textbf{Coherency} & \textbf{\#Instances}\\
\hline
bert-base-uncased         &   10.78 &        2919 \\
bert-large-uncased        &   10.06 &        2919 \\
InformBERT                &    9.79 &        2919 \\
t5-base                   &   10.64 &        2919 \\
t5-large                  &   12.45 &        2919 \\
t5-3b                     &   17.39 &        2919 \\
t5-large-ssm              &   16.76 &        2919 \\
t5-3b-ssm                 &   14.57 &        2919 \\
\hline
\multicolumn{3}{c}{Autoregressive PLMs}  \\
\hline
gpt2                    &    2.36 &        2919 \\
gpt-neo-1.3B &    4.89 &        2919 \\
gpt-neo-2.7B &   11.50 &        2919 \\
\hline
\end{tabular}
\caption{Coherency of different PLMs on a subset of one-token entities using BERT's tokenizer with manually-written prompts.}
\label{tab:results_normalized}
\end{table}

\begin{table}[h]
\centering
\begin{tabular}{lll}
\hline
\textbf{Model} & \textbf{Coherency} & \textbf{\#Instances}\\
\hline
bert-base         &   10.78 &        2919 \\
bert-large        &   10.06 &        2919 \\
InformBERT        &    9.79 &        2926 \\
t5-base                   &    9.66 &       29672 \\
t5-large                  &   10.55 &       29672 \\
t5-3b                     &   16.26 &       29672 \\
t5-large-ssm              &   \textbf{17.06} &       29672 \\
t5-3b-ssm                 &   14.92 &       29672 \\
\hline
\multicolumn{3}{c}{Autoregressive PLMs}  \\
\hline
gpt2               &    2.11 &       29672 \\
gpt-neo-1.3B       &    6.65 &       29672 \\
gpt-neo-2.7B       &    6.19 &       29672 \\

\hline
\multicolumn{3}{l}{\textbf{w/ optimized prompts}}  \\
\hline
bert-base   &    \textbf{7.16} &        2919 \\
bert-large  &    4.62 &        2919 \\
\hline
\multicolumn{3}{l}{\textbf{w/ evidence paragraphs}}  \\

\hline
bert-base-uncased   &   31.09 &        2919 \\
bert-large-uncased  &   31.52 &        2919 \\
InformBERT          &   44.74 &        2926 \\
t5-base   & \underline{\textbf{49.85}} &       29672 \\
t5-large            &   43.31 &       29672 \\
t5-3b               &   44.98 &       29672 \\
t5-large-ssm        &   46.78 &       29482 \\
t5-3b-ssm           &   45.17 &       29672 \\
\hline
\end{tabular}

\caption{Coherency for different PLMs using manually-written, optimized prompts and evidence paragraphs. The highest performance under each category is in \textbf{bold}, and the best performance overall is \underline{underlined}.}
\label{tab:results_w_instances}
\end{table}

\section{Examples}
We show examples of several failures from different prompts and categorize these in Table~\ref{tab:examples}. 

\label{app:examples}
\begin{table*}
    \centering
    \begin{tabular}{+p{2.5cm}^p{1.7cm}^p{1.5cm}^p{3.5cm}^p{3.5cm}^l}
    \hline
        \textbf{Type} & \textbf{Model} & \textbf{Relation} & \textbf{Forward} & \textbf{Backward} & \textbf{ID} \\ \hline
        \textbf{Coherent \newline\& Correct} & bert-base-uncased & edmonton, alberta & edmonton is the capital of [MASK] \newline$\rightarrow{}$ alberta & [MASK] is the capital of alberta \newline$\rightarrow{}$ edmonton & 1 \\  \midrule
        \textbf{Coherent \newline\& Incorrect} & t5-large & Brunei, Malay & The official language of Brunei is [MASK] \newline$\rightarrow{}$ Bruneian & The official language of [MASK] is Bruneian \newline$\rightarrow{}$ Brunei & 2 \\ \midrule
        \textbf{Incoherent \newline\& Correct (1st)} & bert-base-uncased & lucknow, urdu & The official language of lucknow is [MASK] \newline$\rightarrow{}$ urdu & The official language of [MASK] is urdu \newline$\rightarrow{}$ maldives & 3 \\ 
        \textbf{} & gpt-neo 2.7B & Topeka, Kansas & Topeka is the capital of [MASK] \newline$\rightarrow{}$ Kansas & Kansas’s capital is [MASK] \newline$\rightarrow{}$ Quebec City & 4 \\ 
        Repetition & informBERT & iPhone, Apple & iPhone is produced by [MASK] \newline$\rightarrow{}$ apple & [MASK] is produced by apple \newline$\rightarrow{}$ apple & 5 \\ \midrule
        \textbf{Incoherent \newline\& Incorrect} & bert-large-uncased & lille, nord & lille is the capital of [MASK] \newline$\rightarrow{}$ france & [MASK] is the capital of france \newline$\rightarrow{}$ lyon & 6 \\  
        Repetition & t5-base & Germany, Berlin & The capital of Germany is [MASK] \newline$\rightarrow{}$ Frankfurt am Main & The capital of [MASK] is Frankfurt am Main \newline$\rightarrow{}$ Frankfurt am Main & 7 \\  
        Pronoun & bert-base-uncased & munich, germany & munich is located in [MASK]\newline$\rightarrow{}$ bavaria & [MASK] is located in bavaria \newline$\rightarrow{}$ it & 8 \\ \hline 
    \end{tabular}
\caption{Examples from different PLMs.}
\label{tab:examples}
\end{table*}

\section{Additional Details on Masked Language Models }
\label{app:mlm}
Masked PLMs are trained to predict one or several tokens given a context. This is considered a generalization of the conventional language modeling objective that predicts the next token based on its left context. BERT~\cite{devlin-etal-2019-bert}, an encoder-only model, was trained using the Maksed Language Modeling (MLM) objective. T5, an encoder-decoder model, was also trained using a variant of the MLM objective in addition to a mixture of supervised tasks. In the Salient Span Masking (SSM) versions of T5, the models are additionally trained by masking only entities to push the model to focus more on these~\cite{guu-etal-2020-realm, roberts-etal-2020-much}. Similarly, \citet{sadeq-etal-2022-informask} leverage pointwise mutual information to mask salient tokens in an unsupervised manner. 
Table~\ref{tab:models} provides an overview of the architecture and the number of parameters for each model. 

\begin{table}
\centering
\begin{tabular}{lll}
\hline
\textbf{Model} & \textbf{\#Parameters} & \textbf{Architecture}\\
\hline
bert-base & 110M & encoder-only \\
bert-large & 345M & encoder-only \\
InformBERT & 110M & encoder-only \\
t5-base & 220M & encoder-decoder \\
t5-large & 770M & encoder-decoder \\
t5-3B & 3B & encoder-decoder \\
t5-11B & 11B & encoder-decoder \\
gpt-2 & 117M & decoder-only \\
gpt-neo 1.3B & 1.3B & decoder-only \\
gpt-neo 2.7B & 2.7B & decoder-only \\

\hline
\end{tabular}
\caption{Models with number of parameters and architectures. SSM variants of t5 have the same number of parameters as their normal counterparts.}
\label{tab:models}
\end{table}

\section{Choice of Datasets}
\label{app:datasets}
The LAMA probe~\cite{petroni-etal-2019-language} has been proposed to assess how much factual knowledge is contained in PLMs. We believe it is suitable for the experiments we conduct, since it consists of (subject, relation, object) triples. This allows us to evaluate, how often PLMs can predict one entity (either the subject or object) given the other entity and the relation. Additionally, LAMA covers 41 relations of different types, which helps us provide a coherency estimate based on all of these relations. See Table~\ref{tab:lama} for an overview.
We also used the ParaRel dataset~\cite{elazar-etal-2021-measuring}. This dataset has been proposed to measure the sensitivity of PLMs to paraphrased prompts with respect to factual knowledge. Similarly, we use ParaRel to investigate how the coherency score is affected by paraphrased prompts. All the datasets we used are in English. 
Additionally, we used the prompts obtained by Autoprompt~\cite{shin-etal-2020-autoprompt} to investigate the effect of having optimized prompts on the performance. We manually create prompts for autoregressive PLMs. These templates are included with our code.~\footnote{\url{https://github.com/paulyoussef/coherency}}

\begin{table}
\centering
\begin{tabular}{lll}
\hline
\textbf{\#Relations} & \textbf{\#Instances} & \textbf{Example}\\
\hline
41 & 29672 & \texttt{X was born in Y} \\
\hline
\end{tabular}
\caption{Statistics of LAMA and an example.}
\label{tab:lama}
\end{table}

\begin{table}
\centering
\begin{tabular}{lll}
\hline
\textbf{Dataset } & \textbf{License} \\
\hline
LAMA & CC-BY-NC 4.0   \\
ParaRel & MIT License  \\
Optimized prompts & Apache License 2.0 \\

\hline
\end{tabular}
\caption{Licenses of the datasets used in this work.}
\label{tab:license}
\end{table}

\section{Computational Resources}
In all of our experiments, we use a NVIDIA A100 GPU with 80GB of memory. Our experiments took roughly 25 GPU days.

\end{document}